# Logical and Inequality Implications for Reducing the Size and Complexity of Quadratic Unconstrained Binary Optimization Problems

Fred Glover
School of Engineering & Science, University of Colorado, Boulder, Colorado 80309, USA
fred.glover@colorado.edu

Mark Lewis*
Missouri Western State University, Saint Joseph, Missouri 64507, USA
mlewis14@missouriwestern.edu

Gary Kochenberger
School of Business, University of Colorado, 1250 14th St., Denver, CO, 80202, USA
gary.kochenberger@ucdenver.edu

*corresponding author

Abstract. The quadratic unconstrained binary optimization (QUBO) problem arises in diverse optimization applications ranging from Ising spin problems to classical problems in graph theory and binary discrete optimization. The use of preprocessing to transform the graph representing the QUBO problem into a smaller equivalent graph is important for improving solution quality and time for both exact and metaheuristic algorithms and is a step towards mapping large scale QUBO to hardware graphs used in quantum annealing computers. In an earlier paper (Lewis and Glover, 2016) a set of rules was introduced that achieved significant QUBO reductions as verified through computational testing. Here this work is extended with additional rules that provide further reductions that succeed in exactly solving 10% of the benchmark QUBO problems. An algorithm and associated data structures to efficiently implement the entire set of rules is detailed and computational experiments are reported that demonstrate their efficacy.

## 1. Introduction

Given a graph G = [N, E] where N = {1, 2, …, i, … n} where $n = |N|$ is the number of nodes in the graph and E = {(i,j): i, j ∈ N, i ≠ j } is the set of ordered pairs of edges (arcs) between nodes *i* and *j*. Denoting the weight of edge (*i, j*) by $c_{ij}$, we define the Quadratic Unconstrained Binary Optimization Problem (QUBO) as:

$$\text{Maximize: } x_o = \sum_{i \in N} c_{ii} x_i + \sum_{(i,j) \in E} c_{ij} x_i x_j \quad \text{subject to } x_i = \{0,1\} \text{ where } i \in N$$

The equivalent compact definition with the coefficients of (1) represented as a $Q$ matrix is:

$$x_o = \text{Max } x^t Q x: \ x \in \{0, 1\}^n$$

where $Q$ is an *n*-by-*n* square symmetric matrix of constant coefficients. We have represented the diagonal coefficients $c_{ii}$ of Q by $c_i$ in (1) for convenience, and observe that the term $x_i c_{ii} x_i$ from the Q matrix representation reduces to $c_i x_i$ since $x_i^2 = x_i$ for a binary variable. As a clarification of terminology, the Q matrix represents the graph G with nodes N having weights $c_i$ on the Q diagonal (linear elements) and edges E having weights $c_{ij}$ on the Q off-diagonal (quadratic elements).

## 2. Literature

QUBO has been extensively studied (see the survey (Kochenberger, et al., 2014)) and is used to model and solve many categories of optimization problems including network flows, scheduling, max-cut, max-clique, vertex cover and other graph and management science problems A major benefit of QUBO is that it provides a unified modeling framework (Kochenberger, et al., 2004) such that one QUBO algorithm applies to many problem types. NP problems such as graph and number partitioning, covering and set packing, satisfiability, matching, spanning tree as well as others can be converted to Ising form as shown in (Lucas, 2014). Ising problems replace $x \in \{0, 1\}^n$ by $x \in \{-1, 1\}^n$ and can be put in the form of (1) by defining $x_{j'} = (x_j + 1)/2$ and then redefining $x_j$ to be $x_{j'}$. Ising problems are often solved with annealing approaches in order to find a lowest energy state.

Although QUBO problems are NP-complete, good solutions to large problems can be found using modern metaheuristics (Glover, et al., 1998). In addition, a new type of quantum computer based on quantum annealing with an integrated physical network structure of qubits known as a Chimera graph has also been demonstrated to very quickly find good solutions to QUBO (Boixo, et al., 2014).

Related previous work on reducing the size of the QUBO problem can be found in the work of (Kennington & Lewis, 2004) who report rules for reducing multi-commodity networks based on the structure of the network. For certain classes of very structured problems such as vertex cover, max-cut and max-clique, the work of (Boros, et al., 2006) shows that complete reduction can be achieved via computation of the roof duals of the associated capacitated implication network in association with rules involving first and second order derivatives. Similarly, maximum flow and multi-commodity flow networks can be used to help determine QUBO optimal variable assignments and lower bounds (Wang & Kleinberg, 2009) (Adams & Dearing, 1994). The paper by (Lewis & Glover, 2016) presents and tests four rules based directly on the structure of the coefficients in the $Q$ matrix, iteratively applying them to reduce the size of the QUBO problem until no further reductions are possible. This work also explores transformations to reduce a node's edge density (with application to hardware graphs such as the Chimera) and discusses applications to sensitivity analysis.

Benchmark QUBO problems are often highly structured, or have uniform distributions, or are dense, or random but not necessarily connected (Pardalos & Rodgers, 1990). Classic problems with wide application such as the maximum cut problem are highly structured, e.g. all quadratic coefficients are negative and all linear coefficients are positive, or quadratic coefficients are -1s and linear coefficients are positive sums of quadratic coefficients. The rules presented here for predetermining optimal values for qualifying variables are applicable to any QUBO problem, but are most applicable to Q matrices having structural characteristics associated with real-world graphs (sometimes called complex networks (Kim & Wilhelm, 2008)).

The remainder of this paper is organized as follows. Section 3 presents the rules for assigning values to single variables and for generating inequalities and equations involving pairs of variables, thus reducing $Q$ both by value assignments and by setting variables equal to other variables or to complements of other variables. Section 4 introduces more complex rules that allow pairs of variables to receive values simultaneously. Special implementation procedures are also identified in Sections 3 and 4 to permit the associated reductions of Q to be executed efficiently. A pseudocode for implementing the rules and observations of the preceding sections is given in Section 5, and Section 6 presents the experimental design factors, test run parameters and analysis of the test results that compare the outcomes of solving QUBO problems using CPLEX with and without our preprocessing rules. Finally, our summary and conclusions are given in Section 7.

**3. Rules for Fixing Variables and Reducing Q**

In this section we present an analysis that yields rules for reducing $Q$ that include the rules of Lewis and Glover (2016) as well as additional rules that allow Q to be reduced more thoroughly. In the process we also identify relationships that give a foundation for more complex rules described in Section 4 that yield further reductions.

## 3.1 Assigning Values to Single Variables

**Key Observation**. The objective function

$$x_o = \sum(c_i x_i: i \in N) + \sum(c_{ij} x_i x_j: i, j \in N, i \neq j) \text{ or } \sum_{i \in N} c_{ii} x_i + \sum_{(i,j) \in E} c_{ij} x_i x_j \quad (1)$$

can be written for a complete graph as

$$x_o = c_i x_i + \sum(c_k x_k: k \in N\setminus\{i\}) + \sum x_k(\sum(c_{kj} x_j: j \in N\setminus\{k\}): k \in N) \quad (2)$$

The last term in (2) can be written as

$$x_i \sum c_{ij} x_j: j \in N\setminus\{i\}) + \sum(x_k(\sum(c_{kj} x_j: j \in N\setminus\{k\})): k \in N\setminus\{i\}) \quad (2.1)$$

and the last term of (2.1) can be written as

$$\sum(x_k(c_{ki} x_i + \sum(c_{kj} x_j: j \in N\setminus\{k,i\}): k \in N\setminus\{i\})$$
$$= x_i \sum(c_{ki} x_k: k \in N\setminus\{i\}) + \sum(x_k \sum(c_{kj} x_j: j \in N\setminus\{k,i\}): k \in N\setminus\{i\}). \quad (3)$$

Collecting terms from (2), (2.1) and (3) containing $x_i$, writing the 1$^{st}$ term of (3) as $x_i \sum c_{ji} x_j: j \in N\setminus\{i\})$, and defining $d_{ij} = c_{ij} + c_{ji}$, enables us to write

$$x_o(x_i) = x_i(c_i + \sum(d_{ij} x_j: j \in N\setminus\{i\})) \quad (A1)$$

The residual terms that exclude $x_i$ from (2) and (3) can be combined to give

$$x_o(N\setminus\{i\}) = \sum(c_k x_k: k \in N\setminus\{i\}) + \sum(x_k \sum(c_{kj} x_j: j \in N\setminus\{k,i\}): k \in N\setminus\{i\}) \quad (B1)$$

and hence

$$x_o = x_o(x_i) + x_o(N\setminus\{i\}).$$

We write (A1) as

$$x_o(x_i) = x_i V(x_i) \text{ where } V(x_i) = c_i + \sum(d_{ij} x_j: j \in N\setminus\{i\}) \quad (A1°)$$

*Implementation Remark 1*: A value $c_o$ is maintained to identify the amount added to the objective function, where $c_o$ is initialized at 0 before making any changes. Then setting $x_i = 1$ results in $c_o := c_o + c_i$, and also causes Q to be updated by setting $c_j := c_j + d_{ij}$ for all $j \in N\setminus\{i\}$. This update results from the fact that $c_{ij} x_i x_j$ and $c_{ji} x_j x_i$ are replaced respectively by $c_{ij} x_j$ and $c_{ji} x_j$, hence adding $d_{ij} = c_{ij} + c_{ji}$ to $c_j$. After this update, N itself changes by setting $N := N\setminus\{i\}$, and consequently each future update of Q can be limited to elements of the current N. The set N is similarly updated to become N := N\{i} when $x_i$ is assigned a value of 0, though without updating either $c_o$ or Q.

Note we assume that N is always updated as in Implementation Remark 1 so that N does not contain the index of any variable that has been assigned a value. Hence the stipulation k ∈ N for some index k implies that $x_k$ has not been previously set to 0 or 1 by any of the rules given below. More generally, throughout the following, any index mentioned will always be assumed to belong to the current index set N.

For a given value v = 0 or 1, we say $x_i$ = *v is optimal* if $x_i$ = v in some optimal QUBO solution, and say $x_i$ = *v is uniquely optimal* if $x_i$ = v in all optimal QUBO solutions.

Define $Min(V(x_i))$ $(Max(V(x_i)))$ to be a lower (upper) bound on the value of $V(x_i)$ that will maximize $x_o(x_i)$. Since $x_i \geq 0$, we also have $x_o(x_i) \geq x_i Min(V(x_i))$ and $x_o(x_i) \leq x_i Max(V(x_i))$
In the following, we make repeated use of the following results:

**Lemma 1.0**: If $Min(V(x_i)) \geq 0$, then $x_i = 1$ is optimal, and if $Min(V(x_i)) > 0$, then $x_i = 1$ is uniquely optimal.

*Proof*: By (A1°), for $x_i = 0$, $x_o(x_i) = 0$ and for $x_i = 1$, $x_o(x_i) = V(x_i) \geq Min(V(x_i))$. Hence $x_i = 1$ is optimal if $Min(V(x_i)) \geq 0$ and uniquely optimal if $Min(V(x_i)) > 0$.

**Lemma 2.0:** If $Max(V(x_i)) \leq 0$, then $x_i = 0$ is optimal, and if $Max(V(x_i)) < 0$ then $x_i = 0$ is uniquely optimal.

*Proof*: Again by (A1°), for $x_i = 0$, $x_o(x_i) = 0$ and for $x_i = 1$, $x_o(x_i) \leq Max(V(x_i))$. Hence $x_i = 0$ is optimal if $Max(V(x_i)) \leq 0$ and uniquely optimal if $Max(V(x_i)) < 0$.

Let $D_i^-$ and $D_i^+$ respectively denote the sum of the negative $d_{ij}$ values and the positive $d_{ij}$ values over j ∈ N\{i}, which we express by

$$D_i^- = \sum(d_{ij}: d_{ij} < 0, j \in N\setminus\{i\}) \text{ and } D_i^+ = \sum(d_{ij}: d_{ij} > 0, j \in N\setminus\{i\}).$$

***Rule 1.0***: If $c_i + D_i^- \geq 0$ (> 0), then $x_i = 1$ is optimal (uniquely optimal).

*Proof*: From the definition $V(x_i) = c_i + \sum(d_{ij}x_j: j \in N\setminus\{i\})$, by setting each $x_j$, j ∈ N\{i\}) so that $x_j = 0$ for $d_{ij} > 0$ and $x_j = 1$ for $d_{ij} < 0$, it follows that $c_i + D_i^- = Min(V(x_i))$. Then Rule 1.0 follows from Lemma 1.0.

***Rule 2.0***: If $c_i + D_i^+ \leq 0$ ($< 0$), then $x_i = 0$ is optimal (uniquely optimal).

*Proof*: Again drawing on the definition $V(x_i) = c_i + \sum(d_{ij}x_j: j \in N\setminus\{i\})$, by setting each $x_j$, $j \in N\setminus\{i\})$, so that $x_j = 1$ for $d_{ij} > 0$ and $x_j = 0$ for $d_{ij} < 0$, it follows that $c_i + D_i^+ = \text{Max}(V(x_i))$, and the conclusion of Rule 2.0 follows from Lemma 2.0.

Note that the inequality $c_i + D_i^- \geq 0$ in Rule 1.0 implies $c_i \geq 0$, and the inequality $c_i + D_i^+ \leq 0$ in Rule 2.0 implies $c_i \leq 0$ (and these implications also hold by replacing "$\geq$" with "$>$" and by replacing "$\leq$" with "$<$").

*Implementation Remark 2*: Once the values $D_i^-$ and $D_i^+$ have been computed, they can be updated as follows to avoid recomputing them. At the same time the updates of Implementation Remark 1 are performed, when $x_i$ is assigned a value of 0 or 1 then $D_j^-$ and $D_j^+$ are updated for $j \in N\setminus\{i\}$ by setting $D_j^- := D_j^- - d_{ij}$ for $d_{ij} < 0$ and $D_j^+ := D_j^+ - d_{ij}$ for $d_{ij} > 0$.

### 3.1.2 Basic rules from combining implications from two variables $x_i$ and $x_h$

For convenience, we write $V(x_i)$ in the form
$$V(x_i) = c_i + d_{ih}x_h + \sum(d_{ij}x_j: j \in N\setminus\{i,h\}) \tag{A1:h}$$

Throughout the following, we write $d_{hi}$ as $d_{ih}$, since these values are the same.

**Lemma 1.1**: If $\text{Min}(V(x_i: x_h)) \geq 0$, then $x_i \geq x_h$ in some optimal QUBO solution and if $\text{Min}(V(x_i: x_h)) > 0$, then $x_i \geq x_h$ in all optimal QUBO solutions.

*Proof*: If $x_h = 0$ is optimal, then $x_i \geq x_h$ for any value of $x_i$. If instead $x_h = 1$, then the inequality $\text{Min}(V(x_i: x_h = 1)) \geq 0$ ($> 0$) implies $x_i = 1$ is (uniquely) optimal by Lemma 1.0, and hence $x_i \geq x_h$ in some (all) optimal QUBO solution(s).

**Lemma 2.1:** If $\text{Max}(V(x_i): x_h = 1) \leq 0$, then $x_i + x_h \leq 1$ in some optimal QUBO solution and if $\text{Max}(V(x_i): x_h = 1) < 0$, then $x_i + x_h \leq 1$ in all optimal QUBO solutions.

*Proof*: If $x_h = 0$ is optimal, then $x_i + x_h \leq 1$ for any value of $x_i$. If instead $x_h = 1$, the inequality $\text{Max}(V(x_i: x_h = 1)) \leq 0$ ($< 0$) implies $x_i = 0$ is (uniquely) optimal by Lemma 2.0 and hence "$x_i + x_h \leq 1$" in some (all) optimal QUBO solution(s).

To apply these results, we note that when $x_h = 1$, (A1:h) implies
$$V(x_i: x_h = 1) = c_i + d_{ih} + \sum(d_{ij}x_j: j \in N\setminus\{i,h\}) \tag{A1:$x_h$=1}$$

For all of the rules that follow, we assume Rules 1.0 and 2.0 do not assign a value to either $x_i$ or $x_h$ (when the latter takes the role of $x_i$). We will first state these rules separately, in order to justify their respective conclusions. Afterward, in the next subsection we summarize all of the results that yield the same conclusion and also indicate how to enforce the inequalities implied by these rules by weighting certain coefficients appropriately.

***Rule 1.1***. Assume $d_{ih} > 0$. If $c_i + d_{ih} + D_i^- \geq 0$, then $x_i \geq x_h$ in some optimal QUBO solution and if $c_i + d_{ih} + D_i^- > 0$, then $x_i \geq x_h$ in all optimal QUBO solutions.

*Proof*: From (A1:$x_h$=1) it follows that a legitimate value for Min(V($x_i$): $x_h = 1$) is $c_i + d_{ih} + \sum(d_{ij}: d_{ij} < 0, j \in N\setminus\{i,h\}) = c_i + d_{ih} + D_i^-$ since $d_{ih} > 0$ implies $D_i^- = \sum(d_{ij}: d_{ij} < 0: j \in N\setminus\{i,h\})$. Then Rule 1.1 is a direct consequence of Lemma 1.1.

Rule 1.1 is also valid for $d_{ih} = 0$, but this is an uninteresting case since then Rule 1.1 is dominated by Rule 1.0, which yields $x_i = 1$.

***Rule 2.1***. Assume $d_{ih} < 0$. If $c_i + d_{ih} + D_i^+ \leq 0$, then $x_i + x_h \leq 1$ in some optimal QUBO solution and if $c_i + d_{ih} + D_i^+ < 0$ then $x_i + x_h \leq 1$ in all optimal QUBO solutions.

*Proof*: From (A1:$x_h$=1) it follows that a legitimate value for Max(V($x_i$): $x_h = 1$) is $c_i + d_{ih} + \sum(d_{ij}: d_{ij} > 0, j \in N\setminus\{i,h\}) = c_i + d_{ih} + D_i^+$ since $d_{ih} < 0$ implies $D_i^+ = \sum(d_{ij}: d_{ij} > 0: j \in N\setminus\{i,h\})$. Then Rule 2.1 is a direct consequence of Lemma 2.1.

As in the case of Rule 1.1, Rule 2.1 is also valid for $d_{ih} = 0$, but again this is uninteresting since then Rule 2.1 is dominated by Rule 2.0 which implies $x_i = 0$.

Rule 1.1 and Rule 2.1 can give different rules by interchanging the indexes i and h as follows:

***Rule 1.1'***. Assume $d_{ih} > 0$. If $c_h + d_{ih} + D_h^- \geq 0$, then $x_h \geq x_i$ in some optimal QUBO solution and if $c_h + d_{ih} + D_h^- > 0$, then $x_h \geq x_i$ in all optimal QUBO solutions.

***Rule 2.1'***. Assume $d_{ih} < 0$. If $c_h + d_{ih} + D_h^+ \leq 0$, then $x_i + x_h \leq 1$ in some optimal QUBO solution and if $c_h + d_{ih} + D_h^+ < 0$ then $x_i + x_h \leq 1$ in all optimal QUBO solutions.

It is interesting that Rule 1.1' yields a different conclusion from Rule 1.1, but Rule 2.1' yields the same conclusion as Rule 2.1 under different assumptions.

### 3.1.3 Rules Obtained by Complementing Variables

The foregoing rules can be extended to produce additional rules by complementing one or both of $x_i$ and $x_h$, i.e., replacing $x_i$ by $1 - y_i$ and/or replacing $x_h$ by $1 - y_h$, for the complementary 0-1 variables $y_i$ and $y_h$.

These rules rest on identifying the form of $V(x_i: x_h = 1)$ that arises in each of these cases. In particular, we use the notation $V(x_i: y_h)$ to correspond to the case of complementing $x_h$, $V(y_i: x_h)$ to correspond to the case of complementing $x_i$, and $V(y_i: y_h)$ to correspond to the case of complementing both $x_i$ and $x_h$. (By this notation, $V(x_i: x_h)$ would therefore be the same as $V(x_i)$.)

We start from:
$$x_o(x_i) = x_i V(x_i), \text{ where}$$
$$V(x_i) = c_i + d_{ih}x_h + \sum(d_{ij}x_j: j \in N\setminus\{i,h\}).$$

Hereafter, for simplicity, we will only state the rules for the case where an inequality holds in some optimal QUBO solution, since the case where the inequality holds in all optimal QUBO solutions is evident.

### Case 1: For Complementing $x_h$

$x_o(x_i) = x_i V(x_i, y_h)$, where $V(x_i: y_h) = c_i + d_{ih}(1 - y_h) + \sum(d_{ij}x_j: j \in N\setminus\{i,h\})$. Hence we obtain

$V(x_i: y_h = 1) = c_i + \sum(d_{ij}x_j: j \in N\setminus\{i,h\})$ and we can legitimately take

$$\text{Min}(V(x_i: y_h = 1)) = c_i + \sum(d_{ij}: d_{ij} < 0: j \in N\setminus\{i,h\})$$
$$= c_i - d_{ih} + D_i^- \text{ if } d_{ih} < 0$$
$$\text{Max}(V(x_i: y_h = 1)) = c_i + \sum(d_{ij}: d_{ij} > 0: j \in N\setminus\{i,h\})$$
$$= c_i - d_{ih} + D_i^+ \text{ if } d_{ih} > 0$$

Employing the form of Lemmas 1.1 and 2.1 that apply to $V(x_i: y_h = 1)$, we therefore obtain the following results, noting that $x_i \geq y_h$ is the same as $x_i + x_h \geq 1$, and $x_i + y_h \leq 1$ is the same as $x_i \leq x_h$.

***Rule 1.2***. Assume $d_{ih} < 0$. If $c_i - d_{ih} + D_i^- \geq 0$, then $x_i + x_h \geq 1$ in some optimal QUBO solution.
***Rule 2.2***. Assume $d_{ih} > 0$. If $c_i - d_{ih} + D_i^+ \leq 0$, then $x_i \leq x_h$ in some optimal QUBO solution.

The corresponding rules by interchanging the indexes i and h are:

***Rule 1.2'***. Assume $d_{ih} < 0$. If $c_h - d_{ih} + D_h^- \geq 0$, then $x_i + x_h \geq 1$ in some optimal QUBO solution.
***Rule 2.2'***. Assume $d_{ih} > 0$. If $c_h - d_{ih} + D_h^+ \leq 0$, then $x_h \leq x_i$ in some optimal QUBO solution.

### 3.1.4 Implementing the inequalities implied by the rules.

We now observe how the inequalities generated by the foregoing rules can be implemented. If a general 0-1 optimization method is used to solve the QUBO problem, each of the inequalities implied by the foregoing rules can be recorded to be added later to the problem constraints. However, for our present purposes, and for the case where a specialized algorithm is used to solve the QUBO problem, these inequalities can be exploited by identifying an appropriately large value of M and weighting one of the terms $x_i x_h$ or $y_i x_h$ or $x_i y_h$ or $y_i y_h$ by $-M$ to compel the associated product to be 0. In particular, as observed in Kochenberger et al. (2004), the inequalities can be handled as follows.

**For $x_i + x_h \leq 1$**: replace $d_{ih}$ and $d_{hi}$ by $-M$.

**For $x_i \geq x_h$**: replace $c_h$ by $-M$ and $d_{ih}$ and $d_{hi}$ by M. (This results by noting $x_i \geq x_h$ is the same as $y_i + x_h \leq 1$ and hence equivalent to $y_i x_h = 0$. Finally, $y_i x_h = (1 - x_i)x_h = x_h - x_i x_h$,)

**For $x_h \geq x_i$**: replace $c_i$ by $-M$ and $d_{ih}$ and $d_{hi}$ by M. (Interchange the indexes i and h in the prescription for $x_i \geq x_h$.)

**For $x_i + x_h \geq 1$**: replace both $c_i$ and $c_h$ by M and replace $d_{ih}$ by $-M$. (This results by noting $x_i + x_h \geq 1$ is the same as $y_i + y_h \leq 1$ and hence equivalent to $y_i y_h = 0$. Finally $y_i y_h = (1 - x_i)(1 - x_h) = 1 - x_i - x_h + x_i x_h$. Note this adds the constant M to the expression for $x_o$.)

### 3.1.5 Comparisons of Basic Rules in combination to yield additional implications

We must exclude all combinations where $d_{ih} > 0$ in one rule and $d_{ih} < 0$ in the other. Hence the rules for $x_i + x_h \geq 1$ and $x_i + x_h \leq 1$ cannot be combined with the rules for $x_i \leq x_h$ and $x_h \leq x_i$. All other combinations work, to give either $x_i + x_h = 1$ or $x_i = x_h$, as shown next. In all cases we assume neither Rule 1.0 nor Rule 2.0 assigns a value to a variable. Rules that work in combination to give implications:

**Combining rules for $d_{ih} < 0$ that imply $x_i + x_h \geq 1$ and $x_i + x_h \leq 1$: hence $x_i + x_h = 1$**

*Rules that imply $x_i + x_h \geq 1$*

**Rule 1.2**. Assume $d_{ih} < 0$. If $c_i - d_{ih} + D_i^- \geq 0$, then $x_i + x_h \geq 1$ in some optimal QUBO solution.
   $M > Min(0, - (c_i + D_i^-))$.

**Rule 1.2'**. Assume $d_{ih} < 0$. If $c_h - d_{ih} + D_h^- \geq 0$ then $x_i + x_h \geq 1$ in some optimal QUBO solution.
   $M > Min(0, - (c_h + D_h^-))$.

*Rules that imply $x_i + x_h \leq 1$*

**Rule 2.1**. Assume $d_{ih} < 0$. If $c_i + d_{ih} + D_i^+ \leq 0$, then $x_i + x_h \leq 1$ in some optimal QUBO solution.
$M > \text{Min}(0, c_i + D_i^+)$

**Rule 2.1'**. Assume $d_{ih} < 0$. If $c_h + d_{ih} + D_h^+ \leq 0$, then $x_i + x_h \leq 1$ in some optimal QUBO solution.
$M > \text{Min}(0, c_h + D_h^+)$

Similarly, we can identify implications for combining the rules for which $d_{ih} > 0$.

**Combining rules for $d_{ih} > 0$ that imply $x_i \leq x_h$ and $x_i \geq x_h$: hence $x_i = x_h$**

*Rules that imply $x_i \leq x_h$*

**Rule 1.1'**. Assume $d_{ih} > 0$. If $c_h + d_{ih} + D_h^- \geq 0$, then $x_i \leq x_h$ in some optimal QUBO solution.
$M > \text{Min}(0, -(c_h + D_h^-))$.

**Rule 2.2**. Assume $d_{ih} > 0$. If $c_i - d_{ih} + D_i^+ \leq 0$, then $x_i \leq x_h$ in some optimal QUBO solution.
$M > \text{Min}(0, c_i + D_i^+)$.

*Rules that imply $x_i \geq x_h$*

**Rule 1.1**. Assume $d_{ih} > 0$. If $c_i + d_{ih} + D_i^- \geq 0$, then $x_h \leq x_i$ in some optimal QUBO solution.
$M > \text{Min}(0, -(c_i + D_i^-))$.

**Rule 2.2'**. Assume $d_{ih} > 0$. If $c_h - d_{ih} + D_h^+ \leq 0$, then $x_h \leq x_i$ in some optimal QUBO solution.
$M > \text{Min}(0, c_h + D_h^+)$.

From these combinations we respectively obtain the following two substitution rules:

**Rule 2.5**: Assume $d_{ih} < 0$.
If $c_i - d_{ih} + D_i^- \geq 0$ or $c_h - d_{ih} + D_h^- \geq 0$ and if $c_i + d_{ih} + D_i^+ \leq 0$ or $c_h + d_{ih} + D_h^+ \leq 0$
then $x_i + x_h = 1$ in some optimal QUBO solution.

**Rule 2.6**: Assume $d_{ih} > 0$.
If $c_i - d_{ih} + D_i^+ \leq 0$ or $c_h + d_{ih} + D_h^- \geq 0$ and if $c_i + d_{ih} + D_i^- \geq 0$ or $c_h - d_{ih} + D_h^+ \leq 0$
then $x_i = x_h$ in some optimal QUBO solution.

These rules have the novel property that they yield the same conclusions and embody the same conditions (in a different order) when the indexes i and h are interchanged. In other words, if these rules are checked for a given index i and index h, they do not have to be checked again with the indexes i and h interchanged. Together with the use of Implementation Remarks 3 and 4 presented in Appendix C, this property saves additional computation in applying Rules 2.5 and 2.6. However, it is possible to exploit these rules in a even more efficient manners as described by implementation remarks 5 – 7 in the Appendix C.

## 4. Rules for Assigning Values to Pairs of Variables

The rules for assigning values to pairs of variables are more complex than the preceding rules and require more elaborate logic to justify. We again specify the associated lower bounds on M that will yield the outcome specified by each rule, according to the replacements specified in Section 3.1.4, without the need to justify these bounds. In this case, it is sufficient to bound M only according to the inequality associated with the rule, and the stronger outcome that dominates the inequality by assigning specific values to $x_i$ and $x_h$ will hold automatically. We emphasize that making reference to M in these cases is not relevant in typical preprocessing applications, since the assignment of specific values to $x_i$ and $x_h$, removes these variables from further consideration. However, for purposes of sensitivity analysis, it may be useful to know admissible values for M that can produce the outcomes of these rules.

We assume in each case that Rule 1.0 and Rule 2.0 have been applied first, and neither provides the conclusion that $x_i$ or $x_h$ can be assigned a value of 0 or 1.

***Rule 3.1:*** Assume $d_{ih} \geq 0$. If
$$c_i + c_h - d_{ih} + D_i^+ + D_h^+ \leq 0$$
then $x_i + x_h \leq 1$ and moreover $x_i = x_h = 0$ in an optimal QUBO solution.
$$M > \text{Min}(0, c_i + c_h + D_i^+ + D_h^+)$$

*Proof*: First we show $x_i + x_h \leq 1$. Suppose on the contrary that $x_i = x_h = 1$. For any values of $x_i$ and $x_h$ we have $x_o(x_i, x_h) \leq x_o(x_i) + x_o(x_h) = x_iV(x_i) + x_hV(x_h)$, where $x_o(x_i, x_h)$ = the contribution to $x_o$ produced by $x_i$ and $x_h$ together. But the right hand side double counts $d_{ih}x_ix_h$ because $x_o$ only contains $d_{ih}x_ix_h$ once (since $d_{ih} = c_{ih} + c_{hi}$). Removing one instance of the double counted term gives $x_o(i,h) \leq x_iV(x_i) + x_hV(x_h) - d_{ih}x_ix_h \leq x_i\text{Max}(V(x_i)) + x_h\text{Max}(V(x_h)) - d_{ih}x_ix_h$. Setting $x_i = x_h = 1$ gives $x_o(i,h) \leq c_i + c_h + D_i^+ + D_h^+ - d_{ih}$. Since this quantity is $\leq 0$, we conclude it is impossible for both $x_i = x_h = 1$, and the contradiction yields $x_i + x_h \leq 1$.

Next, $c_i + c_h - d_{ih} + D_i^+ + D_h^+ \leq 0$ implies $c_h - d_{ih} + D_h^+ \leq 0$ since $c_i + D_i^+ \geq 0$ must hold due to the fact that Rule 2.0 does not yield $x_i = 0$. However, the result $x_i + x_h \leq 1$ means either $x_i = 0$ or $x_h = 0$. Suppose $x_h = 0$. Then this removes $d_{ih}$ from $D_h^+$ which causes the new value $D_h^{*+}$ of $D_h^+$ to be $D_h^{*+} = D_h^+ - d_{ih}$. Thus the implication that $c_h - d_{ih} + D_h^+ \leq 0$ yields $c_h + D_h^{+*} \leq 0$, which gives $x_i = 0$ by Rule 2.0, and hence both $x_i$ and $x_h = 0$. If instead we suppose $x_i = 0$, then we obtain $x_h = 0$ by the same logic, so again we conclude both $x_i$ and $x_h = 0$, completing the proof.

The preceding analysis discloses that if either $x_i$ or $x_h$ is set to 0 when the conditions of Rule 3.1 hold, then Rule 2.0 will set the other variable to 0. This has the important implication that for any given variable $x_i$, we only need to identify a single variable $x_h$ such that Rule 3.1 will yield $x_i = x_h = 0$, that is, any index h which yields $c_i + c_h - d_{ih} + D_i^+ + D_h^+ \leq 0$. Then all other variables $x_h$ (for different indexes h) that would be paired by the rule to yield $x_i = x_h = 0$, will appropriately be set to 0 using Rule 2.0. Hence effort can be saved by not examining additional variables $x_h$ for a given variable $x_i$ once it is discovered that the rule will set $x_i = 0$.

It is also possible to do more than this by saving the minimum value MinV(i) of $V(i) = c_h - d_{ih} + D_h^+$ (and the associated index h such that MinV(i) = $c_h - d_{ih} + D_h^+$), which is done for each index i in the process of checking Rule 3.1. Then MinV(i) can be updated with a streamlined calculation analogous to the calculation of Remark 4 when other variables are assigned values or are replaced using Rules 2.5 and 2.6. The minimum value of $c_i + c_h - d_{ih} + D_i^+ + D_h^+$ for applying Rule 3.1 is then obtained by adding $c_i + D_i^+$ to MinV(i), making it unnecessary to examine all variables $x_h$ for each $x_i$ on subsequent passes. This more elaborate way of saving computation can also be used in connection with related observations for pairing variables below.

***Rule 3.2:*** Assume $d_{ih} < 0$.   If   $-c_i + c_h + d_{ih} - D_i^- + D_h^+ \leq 0$

Then $x_i \geq x_h$ and moreover $x_i = 1$ and $x_h = 0$ in an optimal QUBO solution.
$$M > \text{Min}(0, -c_i + c_h - D_i^- + D_h^+)$$

*Proof*: Assume $d_{ih} < 0$ and write the inequality of Rule 3.2 both in its original form
$$-c_i + c_h + d_{ih} - D_i^- + D_h^+ \leq 0 \tag{B}$$

and also in the two forms:
$$c_h + D_h^+ \leq c_i - d_{ih} + D_i^- \tag{B1}$$
$$c_h + d_{ih} + D_h^+ \leq c_i + D_i^- \tag{B2}$$

Note that $x_o(x_i) = x_o(1 - y_i) = (1 - y_i)(c_i + \sum(d_{ij}x_j: j \in N\setminus\{i\})) = C(i) + y_iV(y_i)$ for
$C(i) = c_i + \sum(d_{ij}x_j: j \in N\setminus\{i\}))$ and $V(y_i) = -c_i + \sum(-d_{ij}x_j: j \in N\setminus\{i\})$. Hence $\text{Max}(x_o(1 - y_i)) = C(i) + \text{Max}(y_iV(y_i))$. First we show (B) implies $y_i + x_h \leq 1$. Deny this by supposing $y_i = x_h = 1$. For any values of $y_i$ and $x_h$ we have $x_o(x_i, x_h) \leq x_o(x_i) + x_o(x_h) = C(i) + y_iV(y_i) + x_hV(x_h)$, where $x_o(x_i, x_h) = $ the contribution to $x_o$ produced by $x_i$ and $x_h$ together. Write $V(y_i) = -c_i + (-d_{ih}x_h + \sum(-d_{ij}x_j: j \in N\setminus\{i,h\}))$ and $V(x_h) = c_h + d_{ih}x_i + \sum(d_{hj}x_j: j \in N\setminus\{i,h\}))$.

Hence $y_iV(y_i) + x_hV(x_h) = y_i(-d_{ih}x_h) + x_hd_{ih}(1-y_i) = y_ix_h(-2d_{ih}) + x_hd_{ih} +$. But $y_ix_h(-2d_{ih})$ double counts $-d_{ih}y_ix_h$ because $x_o$ only contains $d_{ih}y_ix_h$ once (since $d_{ih} = c_{ih} + c_{hi}$). Removing one instance of the double counted term gives $y_iV(y_i) + x_hV(x_h) = y_i(-c_i + \sum(-d_{ij}x_j: j \in N\setminus\{i,h\}) + x_h(c_h + \sum(d_{hj}x_j: j \in N\setminus\{i,h\}))) - d_{ih}y_ix_h + x_hd_{ih}$ and setting $y_i = x_h = 1$ gives $-c_i + c_h + \sum(-d_{ij}x_j: j \in N\setminus\{i,h\}) + \sum(d_{hj}x_j: j \in N\setminus\{i,h\}))$. A maximum value of this is $-c_i + c_h + \sum(-d_{ij}x: -d_{ij} > 0: j \in N\setminus\{i,h\}) + \sum(d_{hj}:d_{hj} > 0: j \in N\setminus\{i,h\})) = -c_i + c_h - \sum(d_{ij}x: d_{ij} < 0: j \in N\setminus\{i,h\}) + \sum(d_{hj}:d_{hj} > 0: j \in N\setminus\{i,h\}))$ The assumption $d_{ih} < 0$ allows this to be rewritten as $-\sum(d_{ij}x: d_{ij} < 0: j \in N\setminus\{i,h\}) = -(D_i^- - d_{ih})$ and $\sum(d_{hj}:d_{hj} > 0: j \in N\setminus\{i,h\})) = D_h^+$, hence giving a maximum value of $-c_i + c_h + d_{ih} - D_i^- + D_h^+$. Consequently, assuming (B) true implies that $y_i = x_h = 1$ is impossible, giving $y_i + x_h \leq 1$.

Now we observe that (B1) implies $c_i - d_{ih} + D_i^- \geq 0$ since $c_h + D_h^+ \geq (>) 0$ must hold due to the fact that Rule 2.0 does not yield $x_h = 0$. But then by Rule 1.2 we have $x_i + x_h \geq 1$. Since both $x_h \leq x_i$ and $x_i \geq 1 - x_h$, we conclude $x_i = 1$. Upon making this assignment, we update the problem representation by identifying the new value $c_h^*$ of $c_h$ to be $c_h^* = c_h + d_{ih}$ and setting $N := N\setminus\{i\}$. Hence the updated form of $x_o(x_h)$ is $x_o(x_h) = (c_h + d_{ih})x_h + \sum(d_{hj}x_hx_j: j \in N\setminus\{i,h\}$. Denoting the new value of $D_h^+$ after this update by $D_h^{+*}$, and noting that $d_{ih} < 0$, we have

$$c_h^* + D_h^{+*} = c_h^* + \sum(d_{hj}: d_{hj} > 0: j \in N\setminus\{i,h\}) = c_h + d_{ih} + D_h^+$$

By (B2) we then obtain

$$c_h + d_{ih} + D_h^+ \leq c_i + D_i^-$$

and hence

$$c_h^* + D_h^{+*} \leq c_i + D_i^-$$

Since we assume Rule 1.0 did not give $x_i = 1$ previously, we know $c_i + D_i^- \leq 0$, and thus $c_h^* + D_h^{+*} \leq 0$. Rule 2.0 now establishes $x_h = 0$ is optimal and the proof is complete.

Here, similarly to the case of Rule 3.1, our proof of Rule 3.2 yields the conclusion that given $x_i$, if any variable $x_h$ is identified that Rule 3.2 results in setting $x_i = 1$ and $x_h = 0$, the assignment $x_i = 1$ by itself will assure that all additional variables $x_h$ that can be paired with $x_i$ to satisfy the conditions of Rule 3.2 will receive the assignment $x_h = 0$ by Rule 2.0. Consequently, Rule 3.2 does not need to be checked for additional variables $x_h$ that may be coupled with $x_i$ once the first such variable is found for which Rule 3.2 yields $x_i = 1$ and $x_h = 0$.

***Rule 3.3:*** Assume $d_{ih} < 0$. If
$$c_i - c_h + d_{ih} + D_i^+ - D_h^- \leq 0$$
then $x_h \geq x_i$ and moreover $x_i = 0$ and $x_h = 1$ in an optimal QUBO solution.

$$M > \text{Min}(0, c_i - c_h + D_i^+ - D_h^-)$$

This rule is the same as Rule 3.2 upon interchanging the indexes i and h. This brings us to our final rule of this section.

***Rule 3.4***: Assume $d_{ih} \geq 0$. If
$$-c_i - c_h - d_{ih} - D_i^- - D_h^- \leq 0$$
then $x_i + x_h \geq 1$ and moreover $x_i = x_h = 1$ in an optimal QUBO solution.
$$M > \text{Min}(0, -c_i - c_h - D_i^- - D_h^-)$$

*Proof*: We establish this rule by the device of replacing $x_i$ by $1 - y_i$ and $x_h$ by $1 - y_h$ as in Case 3 in Section 3. This allows Rule 3.4 to be treated as an instance of Rule 3.1, where the conclusion $y_i = y_h = 0$ of this instance of Rule 3.1 thus yields $x_i = x_h = 1$. Specifically, let $c_i(y)$, $c_h(y)$, $d_{ih}(y)$, $D_i^+(y)$ and $D_h^+(y)$ denote the values that result for $c_i$, $c_h$, $d_{ih}$, $D_i^+$ and $D_h^+$ as a result of the substitution $x_i = 1 - y_i$ and $x_h = 1 - y_h$. The previous derivation in Case 3 shows that $c_i(y) = (-c_i - d_{ih})$, $c_h(y) = (-c_h - d_{ih})$, $d_{ih}(y) = d_{ih}$, $D_i^+(y) = d_{ih} - D_i^-$ and $D_h^+(y) = d_{ih} - D_h^-$. Writing Rule 3.1 in terms of these values and combining terms gives the statement:

If $d_{ih} \geq 0$ and $-c_i - c_h - d_{ih} - D_i^- - D_h^- \leq 0$ then $y_i = y_h = 0$ in an optimal QUBO solution.
This corresponds to the statement of Rule 3.4 and hence the proof is complete.

Given the connection between Rule 3.1 and 3.4, we can apply the earlier observation concerning Rule 3.1 to conclude that, given $x_i$, we can discontinue examining variables $x_h$ by Rule 3.4 as soon as the first $x_h$ is found that yields $x_h = x_i = 1$. In this case the remaining variables $x_h$ that qualify to be set to 1 with $x_i$ will be identified by Rule 1.0. A summary of all the rules according to their conclusions as well as associated bounding values M and implementation remarks is provided in Appendix C.

## 5. Algorithm for Implementing the Preprocessing Rules.

We describe an approach for implementing the rules of the preceding sections that has several attractive features. Using the graph orientation, we refer to the elements of N as nodes.

**5.1 Data Structures.** The nodes in N are maintained as an ordered list NList, which begins the same as N by setting NList(i) = i, for i = 1, …, n. As the preprocessing rules subsequently drop nodes i from N as a result of setting $x_i = 0$ or 1, or setting $x_i = x_h$ or $1 - x_h$, or simultaneously setting both $x_i$ and $x_h$ to specified binary values, we remove i and/or h from its position on NList in such a way that we can continue to identify all nodes on the current updated N by an organization that is highly efficient.

For this, the nodes i on NList are accessed by a location index iLoc, where i = NList(iLoc) for iLoc = iLoc1 to iLocEnd. NList is initialized by setting NList(iLoc) = iLoc for iLoc = 1 to n (yielding i = NList(iLoc) for i = 1 to n). As nodes are dropped from NList, the positions of remaining nodes may shift because we write over some of the positions where nodes are dropped (to record the identity of nodes that are not dropped). For example, when node 7 is dropped we may replace it by node 1 (where initially NList(iLoc1) = 1), so that the current assignment NList(7) = 7 is changed to become NList(7) = 1. Simultaneously, iLoc1 = iLoc1 + 1 which in the future avoids accessing the old iLoc1 location where node 1 used to be.

We repeat this process to drop nodes from various positions in a way that makes it possible to apply the rules for assigning values to two variables $x_i$ and $x_h$ efficiently, without the duplication that would result by examining both instances of two nodes i and h represented by the pair (i,h) and the pair (h,i).

In particular, if no nodes are dropped, the process may be viewed as simply looking at each node i in succession from i = 1 to n and examining the nodes h for h < i as partners. To generalize this, NList is divided into two parts, the first consisting of an *h-Group* for the nodes h = NList(hLoc), hLoc = hLoc1 to hLocEnd, and the second consisting of an *i-Group* for the nodes i = NList(iLoc), iLoc = iLoc1 to iLocEnd. An outer loop examines the nodes of the i-Group in succession, and for each such node i, the nodes of the h-Group are examined in succession to create the relevant (i,h) pairs.

The examination of a given node i = NList(iLoc) first involves applying Rules 1.0 and 2.0 to see if $x_i$ can be set to 1 or 0, and if not, then the nodes in the h-Group (h = NList(hLoc), hLoc = hLoc1 to hLocEnd) are examined to apply the Rules, 2.5 and 2.6, and Rules 3.1 to 3.4, that result in dropping $x_i$ and/or $x_h$.

After this process is complete, assuming $x_i$ still is not assigned a value, then node i is moved from the first position in the i-Group to become the last node in the h-Group, by setting hLocEnd = hLocEnd + 1 and NList(hLocEnd) = i. This is followed by setting iLoc = iLoc + 1 to access the next node i = NList(iLoc). Accordingly, the new node i will also be able to be paired with the previous node i which is now a member of the h-Group.

By exploiting this two-group structure properly, we can be assured of always examining all relevant pairs (i,h) without duplication, while still dropping nodes from the i-Group or the h-Group by an assignment that drops $x_i$ or $x_h$.

Specifically, the updating rules for dropping a node from the i-Group or the h-Group are as follows. As can be seen, the rule for dropping node i is extremely simple.

*Dropping node i:* Set iLoc = iLoc + 1. (There is no need in this case to transfer node i to become the last node in the h-Group. Implicitly this operation results in setting iLoc1 = iLoc1 + 1. In fact, throughout this process iLoc1 will be the same as iLoc.)

*Dropping node h*: Let h1 = NList(hLoc1), followed by NList(hLoc) = h1 (writing h over by h1) and then set hLoc1 = hLoc1 + 1.

The step of dropping node h does not have to be followed by setting hLoc = hLoc + 1, as would be done when examining all nodes in the h-group for a given node i. The reason for this is that when node h is dropped, it is always accompanied by dropping node i. Hence upon examining the next node i, the examination of nodes in the h-Group starts over, beginning with h = NList(hLoc1).

*Special case*: There may be only one node in the h-Group, as occurs when hLoc1 = hLocEnd. Then, setting hLoc1 = hLoc1 + 1 will result in hLoc1 = hLocEnd + 1, producing hLoc1 > hLocEnd. This signals an empty h-Group and hence the method automatically avoids examining the h-Group when hLoc1 > hLocEnd.

## 5.2 Basic Algorithm: First Pass

With these preliminaries, we now describe the initial pass of all elements in the i-Group. Afterward, we describe the minor change required in order to carry out subsequent passes of elements in the i-Group when such passes are warranted. We enter the first pass with the NList initialized and the sets $D_i^-$ and $D_i^+$ calculated for all *i* in Q (see Figure 1)

## 5.3 Passes of the Algorithm After the First Pass

Subsequent passes are exactly the same as the First Pass except that they (a) do not incorporate the Initialization step and (b) embed the Main Routine within an outer loop. The outer loop repeats the execution of the Main Routine until verifying that no further variables can be assigned values. This verification relies on making use of EndLoc and NextEndLoc.

Figure 1. Basic Algorithm: First Pass to examine all nodes in the i-Group.

```
While iLoc ≤ iLocEnd  //Go through nodes in the i-Group
    i = NList(iLoc)  // get next node in the list
    Apply Rules 1.0 and 2.0 to x_i
    If x_i = 0 or 1 then
            Perform the Q and D updates of Implementation Remarks 1 and 2
            Update NList
    Else  //Examine elements in the h-Group
        While hLoc ≤ hLocEnd
            h = NList(hLoc)
            If x_h requires re-examination (hLoc < LastExam) then
                Apply Rules 1.0 and 2.0 to x_h
                If x_h = 0 or 1 then
                    Perform the Q and D updates
                    Update NList
            If h has not been dropped from NList then
                Apply the Rules 3.1 to 3.4 for assigning values to x_i and x_h
                If x_i and x_h are assigned values:
                    Perform the updates to Q and D
                Else
                    Apply the Rules 2.5 and 2.6 for replacing x_h
                    If x_h is replaced by 1 – x_i or x_i then
                        Perform Q and D updates
                        Update NList
                    Exit the hLoc loop  // stop checking pairs of variables
            Else  //Increment hLoc to prepare to examine the next h in the h-Group
                hLoc = hLoc + 1// Exit the hLoc Loop  if hLoc > hLocEnd
        Endwhile  //end of the hLoc loop
    Update NList
    If  iLoc > EndLoc then Stop: the Preprocessing is complete
Endwhile
End of First Pass
```

To see how this occurs, notice that the h-Group always starts empty at the beginning of the iLoc loop, and the only way the h-Group gains elements is by transferring nodes to it from the i-Group, which happens when a current node i completes its examination without being dropped. Consequently, once all nodes in the i-Group have been examined by the iLoc loop, they will all have either been dropped or transferred to the h-Group. At this point the h-Group is the set of all surviving nodes, and therefore can become the new i-Group for a new pass that goes through the i-Group elements to see if any additional nodes now qualify to be dropped.

Such an additional pass is warranted only if some node was dropped on the current pass, since if no node was dropped then nothing will have changed and a new pass will not uncover any further changes. On the other hand, the new pass need not examine all the surviving nodes, because any node examined in the i-Group after the last node was dropped will not have a new basis for being dropped unless some node is dropped on a new pass before reaching this node.

The key therefore is to keep track of the last node in the i-Group that is dropped on a given pass, since no node in the i-Group examined after this point needs to be reexamined unless a new node is dropped before reaching it on the next pass. When this node is dropped, the current last element of the h-Group identifies the cutoff point, such that if no node is dropped on the next pass by the time this element is examined, then there is no value in to continuing to look farther.

The variable NextEndLoc identifies the location on NList where this current last element of the h-Goup is found. Consequently, after the pass of the elements in the i-Group is complete, and all surviving elements are in the h-Group, and the h-Group in turn becomes the new i-Group, the location NextEndLoc discloses the position of the last element of the new i-Group that needs to be examined and this location is recorded by setting EndLoc = NextEndLoc, so that the next pass can stop after examining $i$ = NList(EndLoc) (unless some node is dropped first). The pseudo-code for the Main Routine captures this fact in two ways. Whenever a node is dropped (a) NextEndLoc records the last h-Group location and (b) EndLoc is set Large. The outcome of (b) assures that the current pass will not terminate prematurely but will examine all nodes of the i-Group. The outcome of (a) assures that the information for the final node dropped will be used to determine the final value of NextEndLoc, as desired.

We conclude this section by observing that the updating of the $D_j^-$ and $D_j^+$ values as in Implementation Remark 2 (which occurs simultaneously with the update of Q in Implementation Remark 1), can most easily be accomplished by a preliminary examination of the indexes $j \in N$ to initialize $D_i^-$ and $D_i^+$. Then, whenever a variable $x_i$ is assigned a value, the values $D_j^-$ and $D_j^+$ for $j \in N\setminus\{i\}$ are updated by Implementation Remark 2 by setting $D_j^- := D_j^- - d_{ij}$ for $d_{ij} < 0$ and $D_j^+ := D_j^+ - d_{ij}$ for $d_{ij} > 0$ (and the analogous update occurs when $x_h$ is assigned a value). We note that the current elements $j \in N$ for carrying out this update are identified by setting

      $j$ = NList(jLoc), jLoc = hLoc1 to hLocEnd for nodes j in the h-Group
      $j$ = NList(jLoc), jLoc = iLoc to iLocEnd for nodes j in the i-Group

where we refer to iLoc rather than iLoc1 in the i-Group case since iLoc is in fact the current value of iLoc1 which we do not bother to update.

The foregoing algorithm can be terminated before its normal termination rule applies, either by limiting the number of passes or by stopping when the number of changes (of assigning a value to a variable) on a given pass becomes small, on the supposition that few additional changes will be made by executing additional passes.

## 6. Experimental Design and Computational Tests

While the preprocessing rules can in principle be applied to the general QUBO problem instance, our particular interest here was to test the rules on problems with a Chimera type structure of groups of densely connected nodes which are sparsely connected to other groups of densely connected nodes. Accordingly, our test problems were generated with these features in mind. An experimental design approach was used to identify the main Q characteristics affecting efficacy of the preprocessor we named QPRO+. Six Q factors, or characteristics, were considered for their effect on three outputs of interest: percent Q reduction, objective value quality and time to best solution. The factors and their settings used in the experimental design are described in Table 1.

As shown in Table 1, we created a $2^{6-2}$ fractional factorial design resulting in 16 tests for each of the 3 problem sizes and 2 problem densities, creating a total of 96 tests with detailed results provided in the Appendices.

Table 1. Q Factors and their Low / High Settings

| Factor ID | Description | Low | High |
|---|---|---|---|
| 1 | -Upper Bound $< c_{ij} <$ Upper Bound | 10 | 100 |
| 2 | Linear Multipliers | 5 | 10 |
| 3 | Quadratic Multipliers | 10 | 20 |
| 4 | % Quadratic Multiplied | 5 | 15 |
| 5 | % Linear Multiplied | 10 | 20 |
| 6 | % non-zero Linear elements | 5 | 25 |

To create test problems with the desired features, the Q generator wass programmed so that a majority of Q elements are drawn from a bounded uniform distribution. A certain percentage of these elements are subsequently increased so that they may fall outside the limits of uniformity, hence creating outliers. The factors determining the Q characteristics are defined in Table 1. The first Factor 1 in the table sets the range of the uniform random number generator, for example a setting of 10 indicates that the coefficients $c_{ij}$ are uniformly distributed between -10 and +10. The second and third factors are multipliers used to increase randomly chosen diagonal (linear) and off-diagonal (quadratic) elements. These factors work together such that setting factors 2 and 5 to their low setting means that 10% of the linear elements are multiplied by 5, creating fewer and closer outliers than when those same factors are set high. In a like manner, factors 3 and 4 determine the outliers for the off-diagonal elements.

The problem sizes, number of edges specified for the Q generator and average edge density for the six problem sets are provided in Table 2. For each of the 6 problems in Table 2, sixteen problems were generated using the characterstics given in Table 1, yielding 96 problems in the test set. Looking at the first row in Table 3, the Upper limit of 10 means the elements of Q are initially drawn from a uniform distribution between -10 and 10 and the Linear multiplier of 10 and the % Large Linear of 10% indicate that 10% of the elements will be multiplied by 10, creating elements that could range from -100 to 100 depending on the value of the initial $c_{ij}$. Factor 6 set to 25% indicates that about a fourth of the diagonal of Q will have non-zero values.

An additional feature of the problem generator is that it creates edges E to yield a graph that is similar to a Chimera graph; in other words, the edges are uniformly distributed except for about 1% of the nodes which are densely connected. All 96 Q matrices have this feature, thus the 1000 variable problems have 10 densely connected nodes and the 10000 variable problems have 100 densely connected nodes. While the average densities may seem small, they represent up to 50 edges per node. All problems generated are connected graphs. During preprocessing the graph can become disconnected, which creates multiple independently solvable smaller problems and future research will explore how to quickly check for connectedness and leverage it by partitioning the problem.

Table 2. Problem Characteristics

| Problem SetID | Q size | Edges | Density % |
|---|---|---|---|
| 1000L | 1000 | 5000 | 1 |
| 1000H | 1000 | 10000 | 2 |
| 5000L | 5000 | 25000 | 0.2 |
| 5000H | 5000 | 50000 | 0.4 |
| 10000L | 10000 | 100000 | 0.2 |
| 10000H | 10000 | 500000 | 1 |

Table 3. Experimental Design Factors for the 16 Problem Types

| ID | Upper limit | Linear Multiplier | Quadratic Multiplier | % large Quadratic | % large Linear | % non-zero Linear |
|---|---|---|---|---|---|---|
| 1 | 10 | 10 | 20 | 5% | 10% | 25% |
| 2 | 100 | 10 | 20 | 15% | 20% | 25% |
| 3 | 10 | 5 | 20 | 15% | 10% | 5% |
| 4 | 100 | 5 | 20 | 5% | 20% | 5% |
| 5 | 10 | 10 | 10 | 5% | 20% | 5% |
| 6 | 100 | 10 | 10 | 15% | 10% | 5% |
| 7 | 10 | 5 | 10 | 15% | 20% | 25% |
| 8 | 100 | 5 | 10 | 5% | 10% | 25% |
| 9 | 100 | 5 | 10 | 15% | 20% | 5% |
| 10 | 10 | 5 | 10 | 5% | 10% | 5% |
| 11 | 100 | 10 | 10 | 5% | 20% | 25% |
| 12 | 10 | 10 | 10 | 15% | 10% | 25% |
| 13 | 100 | 5 | 20 | 15% | 10% | 25% |
| 14 | 10 | 5 | 20 | 5% | 20% | 25% |
| 15 | 100 | 10 | 20 | 5% | 10% | 5% |
| 16 | 10 | 10 | 20 | 15% | 20% | 5% |

## 6.1 Test Results Using CPLEX and QPRO+

The rules developed in Section 3 that predetermine variables are R1, R2, F2.5, R2.6, R3.1, R3.2, R3.3 and R3.4. These eight rules were implemented based on the framework described in Section 5 and tested using the benchmark test set. The percent Q reduction, total number of reductions, reductions per rule, time and solution improvements, and the effects of the parameters in Tables 1, 2 and 3 are reported along with the count of implicit inequalities detected by the logical inequality rules of Section 3.1.4.

As a baseline for comparison, the 96 problems were first solved by CPLEX using default settings (with the quadratic-to-linear parameter turned off so that the problems were not linearized). The CPLEX optimizer includes a sophisticated pre-processing aggregator tightly integrated with the optimizer's linear programming framework. A head to head comparison of the CPLEX procedure and QPRO+ on the problems tested here shows the superiority of QPRO+. In this testing QPRO+ is first applied to a problem, followed by solving the reduced problem using the same default CPLEX settings. Thus the optimizer is constant but the reduced problems solved by CPLEX differ according to the reductions discovered.

All tests were performed using 64 bit Windows 7 on an 8-core i7 3.4 GHz processor with 16 GB RAM. All CPLEX testing used all 8 cores in parallel and the following time limits were placed according to the sizes of the problems: 1000 nodes / 100 seconds; 5000 nodes / 300 seconds; 10000 nodes / 600 seconds. Many of these problems were challenging for CPLEX to solve. As an example, problem # 15, a 5,000 variable, low density problem was solved by CPLEX in 3.5 hours. The combined approach of QPRO+ and CPLEX found the optimal solution in 200 seconds.

Overall, our preprocessing rules, embedded in QPRO+, were very successful in assigning values to variables and thus reducing the size of the Q matrix and the modified problem instance left to be solved. Across all 96 problems, QPRO+ produced more than a 45% reduction in size for more than half of the problems. For 10 of the problems (six with 1000 variables, two with 5000 variables, and two with 10000 variables), QPRO+ was able to set 100% of the variables, completely solving the problems.

With the exception of two of the 96 problems, QPRO+ outperformed the CPLEX aggregator preprocessor by a wide margin in terms of the number of reductions found. In no case did the CPLEX preprocessor produce a 100% reduction. Table 4 summarizes the comparison of the average number of reductions found by QPRO+ and the commercial preprocessor embedded in CPLEX.

Table 4: Comparing CPLEX preprocessor and QPRO+

| Problem Category | Average CPLEX Reductions | Average QPRO+ reductions |
|---|---|---|
| 1000L | 108 | 564 |
| 1000H | 36 | 422 |
| 5000L | 561 | 2493 |
| 5000H | 137 | 1774 |
| 10000L | 313 | 3690 |
| 10000H | 36 | 2582 |
| Averages | 198 | 1920 |

Figure 2 drills down in Table 4 and shows the relative magnitude of individual reduction improvements of QPRO+ over the CPLEX preprocessor for each of the 96 problems. The vertical scale is logarithmic and the horizontal axis has been sorted from low to high QPRO+ reductions to highlight that the increase in number of reductions is often an order of magnitude. The figure illustrates that on all but 2 problems, QPRO+ found dramatically more reductions. On two 10000 node high density problems CPLEX found a total of 36 reductions versus 12 found by QPRO+, indicating that neither approach was able to find a significant number of reductions on these problems.

Figure 2: The number of QPRO+ reductions dominate those found by default CPLEX

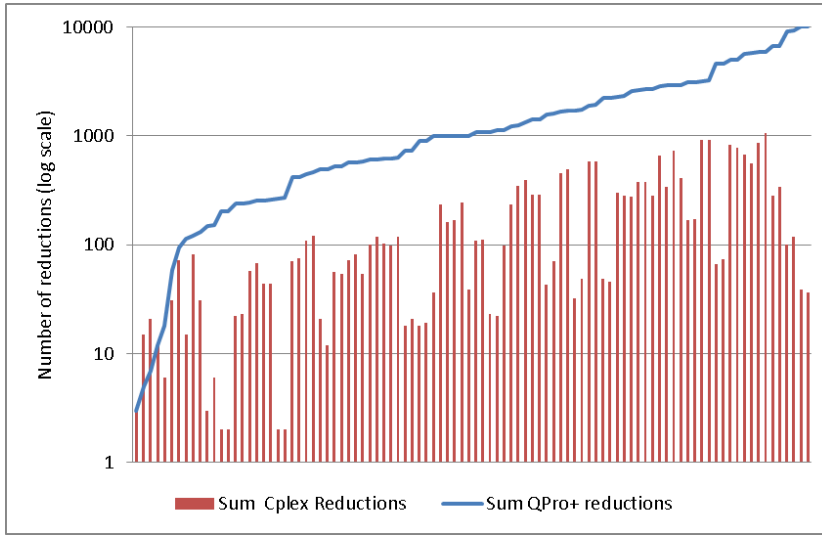

The relationship between problem size, density and average percent reduction is further highlighted in Figure 3 which provides an average percent reduction comparison between QPRO+ and CPLEX's preprocessing procedure, where percent reduction refers to the number of variables eliminated as a percent of the original number. Thus QPRO+ reduced the 1000 variable 2% dense problems by an average of about 45% versus 3% for CPLEX.

Figure 3: Comparing Reductions vs. Density for CPLEX and QPRO+

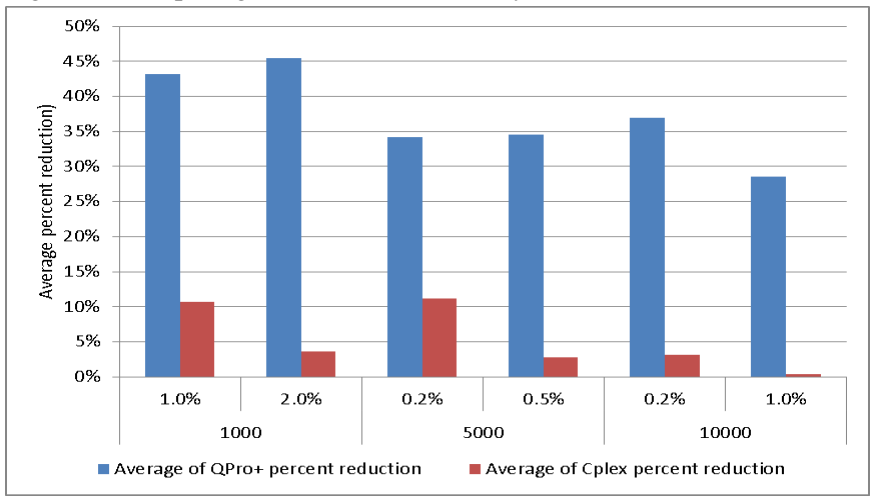

QPRO+ coupled with the CPLEX optimizer was much faster and found better solutions than default CPLEX as summarized in Table 6. For these 96 problems QPRO+ saved about 15000 seconds and found a total improvement of 3082319 in the objective values. The -579 sum of time improvement for the 5000H problems is due to QPRO+ consistently finding better solutions, but not in less time than CPLEX found their best solution.

Table 6: Comparison of Aggregate Solution Performance

| Row Labels | Sum of Time Improvement | Sum of Solution Improvement |
|---|---|---|
| 1000L | 337 | 0 |
| 1000H | 252 | 0 |
| 5000L | 1555 | 8905 |
| 5000H | -579 | 54819 |
| 10000L | 4910 | 2561040 |
| 10000H | 8355 | 457555 |
| Totals | 14830 | 3082319 |

The largest improvements in both time to solution and solution improvement occurred in the larger, denser 10000H problems. For 14 of these 16 problems the combination of QPRO+ and CPLEX was able to find the best solution as its starting incumbent solution and in 2 of these problems QPRO+ found a 100% reduction to yield the optimal solution.

Table 7 summarizes the average percent reductions based on the design factors in Table 1. The table shows that the size of the linear multiplier had the most significant effect on percent reduction because as the multiplier of the linear elements increased from 5x to 10x, the number of reductions on average dropped 27%. Conversely, the effect of multiplying a small percent of the quadratic elements in order to create quadratic outliers, had little effect. This suggests that knowing the structure of Q could be useful in setting expectations for the effectiveness of the progress in a given case.

Table 7. Average Percent Reductions Based on the Design Factors in Table 1

|  | Low Setting | High Setting | Difference |
|---|---|---|---|
| -Upper Bound $< c_{ij} <$ Upper Bound | 44% | 31% | -13% |
| Linear Multipliers | 51% | 24% | -27% |
| Quadratic Multipliers | 36% | 38% | 2% |
| % Quadratic Multiplied | 30% | 45% | 15% |
| % Linear Multiplied | 32% | 42% | 10% |
| % non-zero Linear elements | 42% | 32% | -10% |

Figure 4 shows the decrease in percent reduction as the problem size increases and an exponential increase in preprocessing time. Despite the increase in preprocessing time, note that even for the largest test problems, these times were less than 3.5 seconds. We are currently exploring opportunities to further increase the speed of the preprocessor via more efficient data structures for sparse matrices.

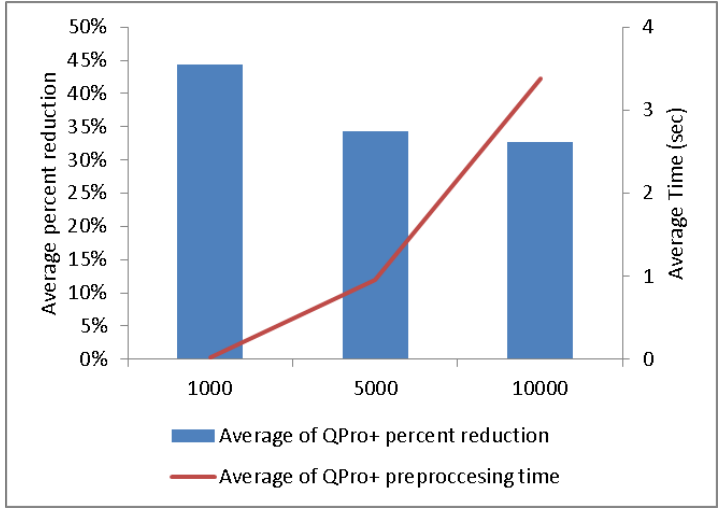

Figure 4: Performance and Time vs Problem Size

The various rules contribute differing amounts to the overall percent reduction (see Table 8). Of the eight rules tested, the two most successful were R1 and R3.4, both of which are concerned with setting variables equal to 1. These two rules accounted for more than 25% of the reductions while. R2 and R3.1, which both set variables equal to 0 together, provided about a 6% reduction. R2.5 and 2.6 based on setting pairs of variables to 0 or 1 together yield about a 5% reduction. However, these rules are not applied individually and when one rule fires it can allow other reductions. We found that the order in which the rules were applied has an effect on the number of reductions achieved. We did not explore alternative orderings in a systematic fashion here but plan to explore this as part of our future work.

Table 8: Percent Reductions by Rule

| Rules | 1000H | 1000L | 5000H | 5000L | 10000H | 10000L | Averages |
| --- | --- | --- | --- | --- | --- | --- | --- |
| R1 | 16.9% | 15.5% | 18.0% | 17.9% | 16.0% | 13.5% | 16.3% |
| R2 | 5.2% | 5.7% | 3.7% | 1.7% | 1.4% | 2.7% | 3.4% |
| R2.5 | 5.1% | 4.3% | 4.4% | 4.4% | 1.4% | 0.2% | 3.3% |
| R2.6 | 1.8% | 1.9% | 1.5% | 0.2% | 2.2% | 3.7% | 1.9% |
| R3.1 | 0.3% | 0.2% | 0.1% | 0.0% | 0.5% | 16.2% | 2.9% |
| R3.2 | 0.6% | 0.6% | 0.4% | 0.1% | 0.1% | 0.0% | 0.3% |
| R3.3 | 0.6% | 0.5% | 0.4% | 0.1% | 0.1% | 0.2% | 0.3% |
| R3.4 | 14.9% | 14.5% | 6.1% | 9.6% | 7.0% | 0.0% | 8.7% |
| Total % avg reduction | 45.5% | 43.2% | 34.6% | 34.2% | 28.5% | 36.5% | |

The majority of reductions found are in the first few passes of the algorithm with subsequent passes finding fewer reductions until none are found and the process terminates. Figure 5 presents the typical percent reductions encountered versus passes of the algorithm. This figure is for the 32 10000 variable problems, but is representative of the other problems tested. The data points represent average percent reductions for 16 problems, hence for the 1% edge dense 10000 variable problems about 600 variables were determined in the first pass (6%) and 2500 on average were determined in the second pass.

Figure 5: Average Percent Reduction Per Pass for the 10000 Variable Problems

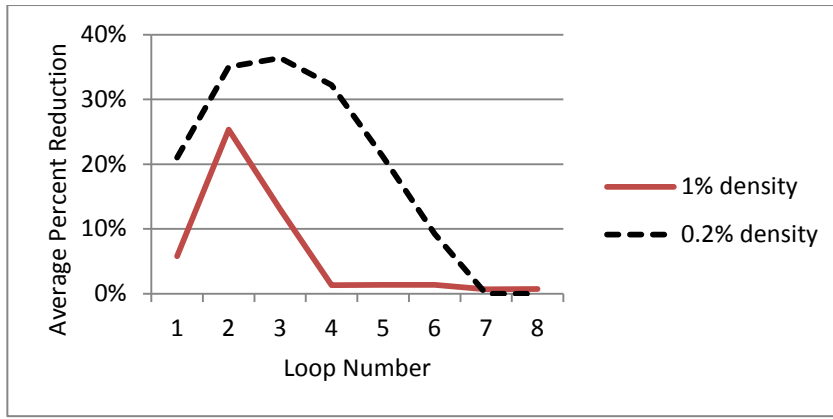

In comparison to our earlier work (Lewis & Glover, 2016) using only 3 basic rules (R1, R2 and R3.4) embedded in a procedure called QPro, we find here that the additional rules yield about a 22% improvement in reduction overall and positively affect every problem. Most importantly the additional rules yield ten 100% reductions in contrast to 0 reductions using only rules 1, 2 and 3.4. The additional rules take slightly longer to run, with the average time for QPRO+ being 1.4 seconds and 0.5 seconds for QPro. A comparison showing the value added in total reductions by the enhanced set of rules over the six sets of problems is summarized in Figure 6.

Figure 6: Improvement in Number of Reductions

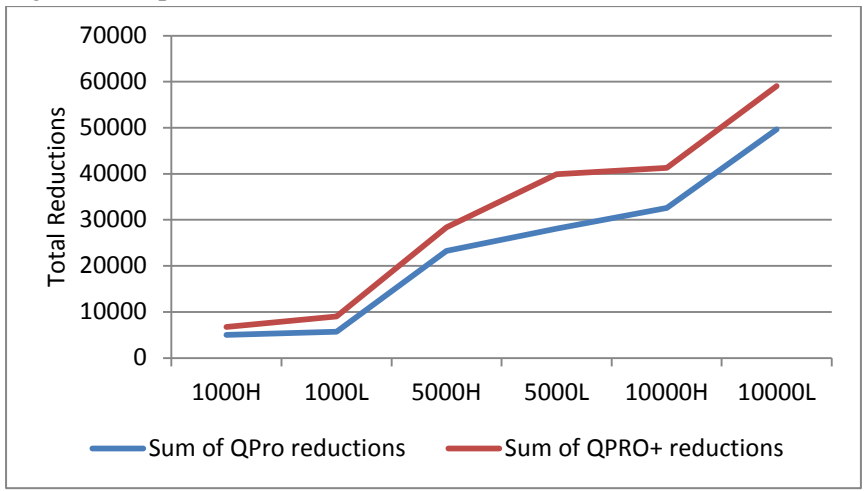

Rules 1.1 through 2.2', are combined into rules 2.5 and 2.6 but individually do not directly set variables, however they can be used individually to discover implicit inequality relationships between pairs of variables.    Figure 7 shows that these rules are effective in discovering a large number of these important inequality relationships.  The problem types that generated many more inequalities than predeterminations were the ones with the lower percentage of quadratic outliers (5% vs. 15%) while the problems with the higher percentage of quadratic outliers produced significantly more reductions. Exploring and leveraging these relationships is another part of our work in progress.

Figure 7: Average number of inequalities generated compared to number of reductions

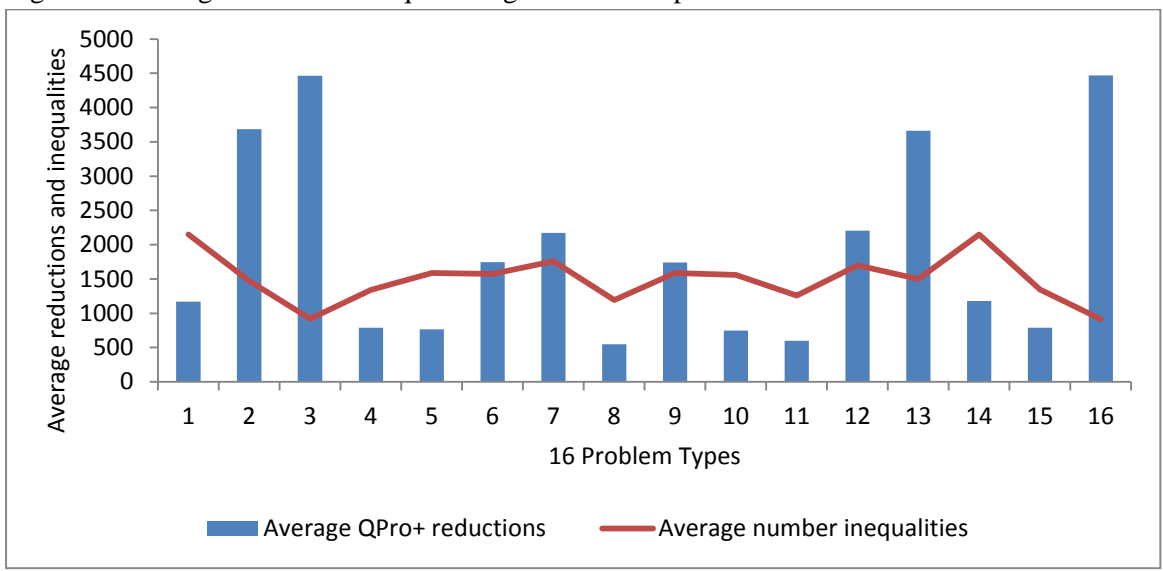

**7.0 Summary and Conclusions**

Our findings demonstrate the value of our preprocessing rules for QUBO problems having structures and densities likely to arise in complex applications. It is noteworthy that the highly refined preprocessing procedure embodied in CPLEX generates an order of magnitude fewer variable settings than our approach

Across the test bed of 96 problems, our rules were successful in setting many variables a priori, leading to significantly smaller problems. In about half of the problems in the test bed QPRO+ achieved a 45% reduction in size and exactly solved 10 problems. The rules also identified many significant implied relationships between pairs of variables resulting in many simple logical inequalities.

Our computational testing based on an implementation of the algorithm detailed herein and using an experimentally designed 96 problem test bed disclosed that (1) sparser problems were more amenable to reduction than denser problems, (2) larger problems required more time to process, (3) a smaller number of linear outliers produced more reductions while changing the magnitude and number of quadratic outliers had little effect, (4) rules 1 and 3.4 accounted for 25% of reductions, and (5) a majority of reductions were made in the first few passes. Investigating the number of inequality relationships between pairs of variables showed that an increase in the number reductions was accompanied by a decrease in the number of inequalities generated.

As part of our on-going research, we are exploring how our results can be specialized to give greater preprocessing reductions in the presence of certain types of additional constraints, including cardinality constraints of the form $\sum(x_j : j \in N) = m$, and multiple choice (GUB) constraints of the form $\sum(x_j : j \in N_r) = 1$, for disjoint subsets $N_r$ of $N$. We intend to report of these and related advances in future papers.

**Appendix A: Efficient Implementation of Preprocessing Rules 2.5 and 2.6**

We first consider how to efficiently implement Rule 2.5 by itself, and then observe how to integrate a corresponding efficient implementation of Rule 2.6 with Rule 2.5.

Write the substitution Rule 2.5 as follows.

***Rule 2.5***: Assume $d_{ih} < 0$.
 (A1) If $c_i - d_{ih} + D_i^- \geq 0$ or (A2) if $c_h - d_{ih} + D_h^- \geq 0$
  and
 (B1) if $c_i + d_{ih} + D_i^+ \leq 0$ or (B2) if $c_h + d_{ih} + D_h^+ \leq 0$
then $x_i + x_h = 1$ in some optimal QUBO solution.

Recall the definitions:
$MaxD_i = Max(d_{ij} > 0: j \in N\setminus\{i\})$ and $MinD_i = Min(d_{ij} < 0: j \in N\setminus\{i\})$.

We say the condition (A1) or (B1) (respectively, (A2) or (B2)) *strongly holds* if it holds when $MinD_i$ (respectively, $MinD_h$) replaces $d_{ih}$ in the statement of Rule 2.5. By our previous observations, if a condition (A1) or (B1) strongly holds then it holds for at least the value $d_{ih}$ such that $d_{ih} = MinD_i$, and if it fails to strongly hold, then it also fails to hold for all $d_{ih}$. The corresponding statement applies to (A2) and (B2) in the case where $d_{ih} = MinD_h$.

When examining each index i in N, define $A(i)$ = True if (A1) strongly holds and $A(i)$ = False otherwise; and similarly define $B(i)$ = True if (B1) strongly holds and $B(i)$ = False otherwise.

We obtain the following useful result. If both $A(i)$ = True and $B(i)$ = True, then we conclude $x_i + x_h = 1$, because (A1) and (B1) hold for the same $d_{ih}$ (= $MinD_i$). On the other hand, if both $A(i)$ = False and $B(i)$ = False (hence both (A1) and (B1) fail to strongly hold) then the only possible way to yield $x_i + x_h = 1$ by Rule 2.5 is if (A2) and (B2) both strongly hold, and this will be discovered by testing whether (A1) and (A2) strongly hold for a different index i that corresponds to the index h.

Consequently, we first apply a *reduced version* of Rule 2.5 by only testing whether (A1) and (B1) strongly hold for each index i. By this approach, we do not have to examine any combinations of indexes i and h, and hence expend no more computational effort than by applying the simple Rules 1.0 and 2.0 for each index i. (Some minor additional effort is required to update $MinD_i$ and $MinD_h$ if an assignment or substitution is made.)

Thus we assume we only apply the reduced version of Rule 2.5 until a complete pass of all indexes i in N yields no more assignments or substitutions. Call this outcome the *early termination*.

Once early termination occurs, we can apply a *residual version* of Rule 2.5 that takes care of all combinations not yet examined in a way that likewise avoids a great deal of computational effort. This residual version exploits the fact that if (A1) strongly holds for i = i1, then (A2) strongly holds for h = i1 when i > i1, and similarly, if (B1) strongly holds for i = i1, then (B2) strongly holds for h = i1 when i > i1. The only cases of interest are when (A1) and (B2) both hold and when (B1) and (A2) both hold.

Save two lists, AList = {i: A(i) = True} and BList = {i: A(i) = True}. A given index i can only be on one of these two lists (since if it is on both, we will have found $x_i + x_h = 1$). Moreover, unless a substitution results following an early termination, no rules need to be tested other than the substitution rules. It is also convenient to keep a list ABList = {i: A(i) = True or B(i) = True} (hence ABList is the union of AList and BList). The only possible combinations of indexes i and h that are relevant to examine are those for i and h both on ABList, where one of these indexes is on AList and the other is on BList. Hence we execute the residual version of Rule 2.5 as follows.

**Residual Version of Rule 2.5**
For each index i on ABList.
    If A(i) = True, then // (A1) holds for at least $d_{ih}$ = $MinD_i$ and we want to check whether
       (A1) holds in combination with some index h such that (B2) also holds.
       For each index h on BList.
          If (A1) and (B2) of Rule 2.5 hold then Rule 2.5 yields $x_i + x_h = 1$ (hence the substitution is executed and the early termination is cancelled, to return to applying the rules with the reduced version of Rule 2.5)
       Endfor
       Remove i from ABList and from AList.
    Else // B(i) = True, and (B1) holds for at least $d_{ih}$ = $MinD_i$ and we want to check whether
       (A2) holds in combination with some index h such that (B1) also holds.
       For each index h on BList.
          If (B1) and (A2) of Rule 2.5 hold then Rule 2.5 yields $x_i + x_h = 1$ (hence the substitution is executed and the early termination is cancelled, to return to applying the rules with the reduced version of Rule 2.5)
       Endfor
       Remove i from ABList and from BList.
    Endif
Endfor

Clearly this method only examines a subset of the index combinations for i and h, which is likely to be a relatively small number compared to all i and h in N.

Now we observe how integrate this treatment of Rule 2.5 with a corresponding treatment of Rule 2.6.

We write Rule 2.6 in the form:

***Rule 2.6***: Assume $d_{ih} > 0$.
    (C1) if $c_i - d_{ih} + D_i^+ \leq 0$ or (C2) If $c_h + d_{ih} + D_h^- \geq 0$ or
           and
    (D1) if $c_i + d_{ih} + D_i^- \geq 0$ or (D2) $c_h - d_{ih} + D_h^+ \leq 0$
then $x_i = x_h$ in some optimal QUBO solution.

Now we define C(i) and D(i) analogously to A(i) and B(i), and the lists CList(i), DList(i) and CDList(i) analogously to AList(i), BList(i) and ABList(i).

The reduced version of Rule 2.6 only tests for each i in N to see whether (C1) and (D1) both strongly hold (which also covers the case where (C2) and (D2) both strongly hold). Once the full preprocessing method terminates with using reduced versions of both Rules 2.5 and 2.6, we apply residual versions of both of these rules. The residual version of Rule 2.6 is then exactly analogous to the residual version of Rule 2.5.

**Appendix B: Alternative Derivations for Logical Inequality Rules**

We demonstrate below that the following two cases yield rules that duplicate the rules previously obtained.

**Case 2: For Complementing $x_i$**

$x_o(x_i) = (1 - y_i)V(x_i, y_h)$, hence $x_o(x_i) = (1 - y_i)(c_i + d_{ih}x_h + \sum(d_{ij}x_j: j \in N\setminus\{i,h\}))$
$= (c_i + d_{ih}x_h + \sum(d_{ij}x_j: j \in N\setminus\{i,h\})) + y_i(-(c_i + d_{ih}x_h + \sum(d_{ij}x_j: j \in N\setminus\{i,h\})))$.

Regardless of the values received by other variables, the only portion of this expression affected by assigning a value to $x_i$, hence to $y_i$, is

$y_i V^o(y_i: x_h)$ where $V^o(y_i: x_h) = -(c_i + d_{ih}x_h + \sum(d_{ij}x_j: j \in N\setminus\{i,h\}))$.

Hence for the analysis of Lemmas 1.1 and 2.1 to apply, we are interested in identifying legitimate values for $\text{Min}(V^o(y_i: x_h = 1))$ and $\text{Max}(V^o(y_i: x_h = 1))$, which are respectively given by

$\text{Min}(V^o(y_i: x_h = 1)) = -(c_i + d_{ih} + \sum(d_{ij}: d_{ij} > 0: j \in N\setminus\{i,h\}))$
$= -(c_i + d_{ih} + D_i^+)$ if $d_{ih} < 0$

$\text{Max}(V^o(y_i: x_h = 1)) = -(c_i + d_{ih} + \sum(d_{ij}: d_{ij} < 0: j \in N\setminus\{i,h\}))$
$= -(c_i + d_{ih} + D_i^-)$ if $d_{ih} > 0$

From this, employing the form of Lemmas 1.1 and 2.1 that apply to $V^o(y_i: x_h = 1)$, and noting that $y_i \geq x_h$ is the same as $x_i + x_h \leq 1$, and $y_i + x_h \leq 1$ is the same as $x_h \leq x_i$, we obtain

***Rule 1.3***. Assume $d_{ih} < 0$. If $c_i + d_{ih} + D_i^+ \leq 0$, then $x_i + x_h \leq 1$ in some optimal QUBO solution.
***Rule 2.3***. Assume $d_{ih} > 0$. If $c_i + d_{ih} + D_i^- \geq 0$, then $x_h \leq x_i$ in some optimal QUBO solution.

The corresponding rules by interchanging the indexes i and h are:

***Rule 1.3'***. Assume $d_{ih} < 0$. If $c_h + d_{ih} + D_h^+ \leq 0$, then $x_i + x_h \leq 1$ in some optimal QUBO solution.
***Rule 2.3'***. Assume $d_{ih} > 0$. If $c_h + d_{ih} + D_h^- \geq 0$, then $x_i \leq x_h$ in some optimal QUBO solution.

**Case 3: For Complementing $x_i$ and $x_h$**

$x_o(x_i) = (1 - y_i)V(y_i, y_h)$, hence $x_o(x_i) = (1 - y_i)(c_i + d_{ih}(1 - y_h) + \sum(d_{ij}x_j: j \in N\setminus\{i,h\}))$
$= (c_i + d_{ih}(1 - y_h) + \sum(d_{ij}x_j: j \in N\setminus\{i,h\})) + y_i(-(c_i + d_{ih}(1 - y_h) + \sum(d_{ij}x_j: j \in N\setminus\{i,h\})))$.

Regardless of the values received by other variables, the only portion of this expression affected by assigning a value to $x_i$, hence to $y_i$, is $y_i V^o(y_i: y_h)$ where $V^o(y_i: y_h) = -(c_i + d_{ih}(1 - y_h) + \sum(d_{ij}x_j: j \in N\backslash\{i,h\}))$. Hence for the analysis of Lemmas 2.1 and 2.2 to apply, we are interested in identifying legitimate values for $Min(V^o(y_i: y_h = 1))$ and $Max(V^o(y_i: y_h = 1))$, which are respectively given by

$Min(V^o(y_i: y_h = 1)) = -(c_i + \sum(d_{ij}: d_{ij} > 0: j \in N\backslash\{i,h\})))$
$\qquad = -(c_i - d_{ih} + D_i^+)$ if $d_{ih} > 0$
$Max(V^o(y_i: y_h = 1)) = -(c_i + \sum(d_{ij}: d_{ij} < 0: j \in N\backslash\{i,h\})))$
$\qquad = -(c_i - d_{ih} + D_i^-)$ if $d_{ih} < 0$

From this, employing the form of Lemmas 1.1 and 2.1 that apply to $V^o(y_i: y_h = 1)$, and noting that $y_i \geq y_h$ is the same as $x_h \geq x_i$ and $y_i + y_h \leq 1$ is the same as $x_i + x_h \geq 1$ we obtain

***Rule 1.4***. Assume $d_{ih} > 0$. If $c_i - d_{ih} + D_i^+ \leq 0$, then $x_h \geq x_i$ in some optimal QUBO solution.
***Rule 2.4***. Assume $d_{ih} < 0$. If $c_i - d_{ih} + D_i^- \geq 0$, then $x_i + x_h \geq 1$ in some optimal QUBO solution.

The corresponding rules by interchanging the indexes i and h are:

***Rule 1.4'***. Assume $d_{ih} > 0$. If $c_h - d_{ih} + D_h^+ \leq 0$, then $x_i \geq x_h$ in some optimal QUBO solution.
***Rule 2.4'***. Assume $d_{ih} < 0$. If $c_h - d_{ih} + D_h^- \geq 0$, then $x_i + x_h \geq 1$ in some optimal QUBO solution.

**Appendix C:   Summarizing the rules according to their conclusions and Implementation Remarks.**

To facilitate comparisons, we now group the rules according to their conclusions. We see that there are a number of duplications. In particular, each of the rules for Cases 2 and 3 above duplicates one of the rules that precede it. Nevertheless, the analysis used in these cases, and particularly in Case 3, proves valuable in establishing later rules for assigning values to pairs of variables. In the following summary, we also indicate a lower bound on the value of M that will insure each rule will imply its associated inequality by the replacement indicated in Section 3.1.4, preceding.

**Rules that imply $x_i + x_h \leq 1$**

***Rule 2.1***. Assume $d_{ih} < 0$. If $c_i + d_{ih} + D_i^+ \leq 0$, then $x_i + x_h \leq 1$ in some optimal QUBO solution.
$\qquad M > Max(0, c_i + D_i^+)$
***Rule 2.1'***. Assume $d_{ih} < 0$. If $c_h + d_{ih} + D_h^+ \leq 0$, then $x_i + x_h \leq 1$ in some optimal QUBO solution.
$\qquad M > Max(0, c_h + D_h^+)$
$\qquad\qquad$**Duplicates of these rules:** *Rule 1.3* and *Rule 1.3' (See Appendix B)*

**Rules that imply $x_i \geq x_h$**

***Rule 1.1***. Assume $d_{ih} > 0$. If $c_i + d_{ih} + D_i^- \geq 0$, then $x_h \leq x_i$ in some optimal QUBO solution.
$\qquad M > Max(0, -(c_i + D_i^-))$.

***Rule 2.2'***. Assume $d_{ih} > 0$. If $c_h - d_{ih} + D_h^+ \leq 0$, then $x_h \leq x_i$ in some optimal QUBO solution.
$M > \text{Max}(0, c_h + D_h^+)$.

**Duplicates of these:** *Rule 2.3*.and *Rule 1.4' (See Appendix B)*

**Rules that imply $x_i \leq x_h$**

***Rule 1.1'***. Assume $d_{ih} > 0$. If $c_h + d_{ih} + D_h^- \geq 0$, then $x_i \leq x_h$ in some optimal QUBO solution.
$M > \text{Max}(0, -(c_h + D_h^-))$.

***Rule 2.2***. Assume $d_{ih} > 0$. If $c_i - d_{ih} + D_i^+ \leq 0$, then $x_i \leq x_h$ in some optimal QUBO solution.
$M > \text{Max}(0, c_i + D_i^+)$.

**Duplicates of these rules:** *Rule 2.3' and Rule 1.4. (See Appendix B)*

**Rules that imply $x_i + x_h \geq 1$**

***Rule 1.2***. Assume $d_{ih} < 0$. If $c_i - d_{ih} + D_i^- \geq 0$, then $x_i + x_h \geq 1$ in some optimal QUBO solution.
$M > \text{Max}(0, -(c_i + D_i^-))$.

***Rule 1.2'***. Assume $d_{ih} < 0$. If $c_h - d_{ih} + D_h^- \geq 0$ then $x_i + x_h \geq 1$ in some optimal QUBO solution.
$M > \text{Max}(0, -(c_h + D_h^-))$.

**Duplicates of these rules:** *Rule 2.4 and Rule 2.4'. (See Appendix B)*

*Implementation Remark 3*: To efficiently implement the foregoing basic rules, it is valuable to initially identify and subsequently update two values associated with each $i \in N$: $\text{MaxD}_i = \text{Max}(d_{ij} > 0: j \in N\setminus\{i\})$ and $\text{MinD}_i = \text{Min}(d_{ij} < 0: j \in N\setminus\{i\})$, along with their associated indexes $\max(i) = \arg\max(d_{ij} > 0: j \in N\setminus\{i\})$ and $\min(i) = \arg\min(d_{ij} < 0: j \in N\setminus\{i\})$. (We set $\text{MaxD}_i = 0$ and $\max(i) = 0$, or $\text{MinD}_i = 0$ and $\min(i) = 0$, respectively, if $\max(i)$ or $\min(i)$ is undefined.) Then each of the foregoing rules can be implemented efficiently from the knowledge of $\text{MaxD}_i$, $\text{MinD}_i$, $\text{MaxD}_h$ or $\text{MinD}_h$ depending on the case. Specifically, Rules 2.1 and 2.1' (where $d_{ih} < 0$) only need to be tested, respectively, for $d_{ih} = \text{MinD}_i$ (hence $h = \min(i)$) and for $d_{ih} = \text{MinD}_h$ (hence for $i = \min(h)$). The same is true for Rules 1.2 and 1.2'. On the other hand, Rules 1.1 and 2.2' (where $d_{ih} > 0$) only need to be tested, respectively, for $d_{ih} = \text{MaxD}_i$ (hence for $h = \max(i)$) and for $d_{ih} = \text{MaxD}_h$ (hence for $i = \max(h)$). The Rules 1.1' and 2.2 reverse the indexes i and h of Rules 1.1 and 2.2', respectively. These restrictions save a great deal of computation.

*Implementation Remark 4*: When a variable $x_i$ is set to 0 or 1, the values $\text{MaxD}_j$ and $\text{MinD}_j$ and the associated indexes $\max(j)$ and $\min(j)$ can be updated for $j \in N\setminus\{i\}$ (once they have been determined initially) at the same time $D_j^-$ and $D_j^+$ are updated by Implementation Remark 2. Specifically, if $d_{ij} > 0$: then if $i \neq \max(j)$ the value $\text{MaxD}_j$ and the index $\max(j)$ are unchanged and otherwise $\text{MaxD}_j$ and $\max(j)$ must be identified from their definitions, $\text{MaxD}_j = \text{Max}(d_{jk} > 0: k \in N\setminus\{j,i\})$ and $\max(j) = \arg\max(d_{jk} > 0: k \in N\setminus\{j,i\})$. (Here the index i is excluded from N in $N\setminus\{j,i\}$) because $x_i$ will be dropped. Alternatively, after dropping $x_i$ and removing i from N, we can refer to $k \in N\setminus\{j\}$ instead of $k \in N\setminus\{j,i\}$.) Similarly, If $d_{ij} < 0$: then if $i \neq \min(j)$ the value $\text{MinD}_j$ and the index $\min(j)$ are unchanged and otherwise must be identified from their definitions, $\text{MinD}_j = \text{Min}(d_{jk} < 0: k \in N\setminus\{j,i\})$ and $\min(j) = \arg\min(d_{jk} < 0: k \in N\setminus\{j,i\})$. We note that Implementation Remarks 3 and 4 can be implemented by storing only the indexes $\max(i)$ and $\min(i)$ for each $i \in N$.

Now we identify the updates of problem coefficients when Rule 2.5 or 2.6 performs the substitution of replacing $x_h$ by $1 - x_i$ or $x_i$.

*Implementation Remark 5*: If $x_i + x_h = 1$, the updates to replace $x_h$ by $1 - x_i$ are
$$N := N\setminus\{h\}$$
$$c_i := c_i - c_h$$
$$d_{ij} := d_{ij} - d_{hj} \text{ for all } j \in N\setminus\{i,h\} \text{ (for } j \in N\setminus\{i\} \text{ after setting } N := N\setminus\{h\})$$
$$c_o := c_o + c_h$$

The foregoing updates are evidently not the same as those that occur by the substitution that replaces $x_i$ by $1 - x$, whose updates are given by interchanging the indexes i and h in Implementation Remark 5.

*Implementation Remark 6*: If $x_i = x_h$, the updates to replace $x_h$ by $x_i$ are:
$$N := N\setminus\{h\}$$
$$c_i := c_h + c_i + d_{ih}.$$
$$d_{ij} := d_{hj} + d_{ij} \text{ for all } j \in N\setminus\{i,h\} \text{ (for } j \in N\setminus\{i\} \text{ after setting } N := N\setminus\{h\})$$

In contrast to Implementation Remark 5, the updates for Implementation Remark 6 are indistinguishable if the indexes i and h are interchanged, except for setting $N := N\setminus\{i\}$ instead of $N := N\setminus\{h\}$.

*Implementation Remark 7*: The values $D_j^-$ and $D_j^+$ can be updated for $j \in N\setminus\{i\}$ at the same time the updates of Implementation Remark 5 or Implementation Remark 6 are performed. Let $d_{ij}$, denote the value of $d_{ij}$ after the update of Remark 5 or 6, and let $d_{ij}'$ denote the value of $d_{ij}$ before executing this update. Then the updated values $D_j^-$ and $D_j^+$ are as follows, for each $j \in N\setminus\{i\}$ (after setting $N := N\setminus\{h\}$): First, if $d_{ij}' < 0$ then set $D_j^- := D_j^- - d_{ij}'$ and otherwise set $D_j^+ := D_j^+ - d_{ij}'$. Then, if $d_{ij} < 0$ set $D_j^- := D_j^- + d_{ij}$ and otherwise set $D_j^+ := D_j^+ + d_{ij}$. (In sum, two "if checks" and one addition and one subtraction are required for each $j \in N\setminus\{i\}$.)